%% file: manuscript.tex
\documentclass[11pt]{article}

\usepackage[margin=1in]{geometry}
\usepackage[utf8]{inputenc}
\usepackage[T1]{fontenc}
\usepackage{amsmath,amssymb,amsfonts}
\usepackage{bbm}
\usepackage{graphicx}
\usepackage{textcomp}
\usepackage{algorithm}
\usepackage{algorithmic}
\usepackage{listings}
\usepackage{tabularx}
\usepackage{import}
\usepackage{svg}
\usepackage{booktabs}
\usepackage{xcolor}
\usepackage{colortbl}
\usepackage{multirow}
\usepackage{makecell}
\usepackage[normalem]{ulem}
\usepackage{caption}
\usepackage[figuresright]{rotating}
\usepackage{hyperref}
\usepackage{url}
\usepackage{nicefrac}
\usepackage{microtype}
\usepackage[numbers,sort&compress]{natbib}
\usepackage{authblk}

\newcommand{\method}{FeatEHR-LLM}

\title{FeatEHR-LLM: Leveraging Large Language Models for Feature Engineering\\in Electronic Health Records}

\author[1]{Hojjat Karami\thanks{Corresponding author. \texttt{hojjat.karami@epfl.ch}}}
\author[1]{David Atienza}
\author[1]{Jean-Philippe Thiran}
\author[1]{Anisoara Ionescu}
\affil[1]{École Polytechnique Fédérale de Lausanne (EPFL), Lausanne, Switzerland}

\date{}

\begin{document}

\maketitle

\begin{abstract}
Feature engineering for Electronic Health Records (EHR) is complicated by irregular observation intervals, variable measurement frequencies, and structural sparsity inherent to clinical time series. Existing automated methods either lack clinical domain awareness or assume clean, regularly sampled inputs, limiting their applicability to real-world EHR data. We present \textbf{FeatEHR-LLM}, a framework that leverages Large Language Models (LLMs) to generate clinically meaningful tabular features from irregularly sampled EHR time series. To limit patient privacy exposure, the LLM operates exclusively on dataset schemas and task descriptions rather than raw patient records. A tool-augmented generation mechanism equips the LLM with specialized routines for querying irregular temporal data, enabling it to produce executable feature-extraction code that explicitly handles uneven observation patterns and informative sparsity. FeatEHR-LLM supports both univariate and multivariate feature generation through an iterative, validation-in-the-loop pipeline. Evaluated on eight clinical prediction tasks across four ICU datasets, our framework achieves the highest mean AUROC on 7 out of 8 tasks, with improvements of up to 6 percentage points over strong baselines. Code is available at \url{github.com/hojjatkarami/FeatEHR-LLM}.
\end{abstract}



\section{Introduction}

Electronic Health Records (EHRs) contain rich longitudinal data that capture patient health trajectories. Within these records, time series measurements, such as vital signs and laboratory results, form the backbone of clinical information. However, modeling this data is challenging because clinical time series are inherently irregularly sampled, manifesting as variable observation frequencies and event-driven measurements. The resulting gaps represent structural sparsity, with measurements recorded only when clinically indicated, rather than statistical missingness \cite{ghassemiReviewChallengesOpportunities2020, zhouMissingDataMatter2023, getzenMiningEquitableHealth2023a}. Despite these complexities, machine learning (ML) is increasingly leveraged to analyze EHRs for critical clinical decision support \cite{chenHarnessingPowerClinical2023,yeRoleArtificialIntelligence2024}.

To effectively utilize this complex temporal data, feature engineering remains fundamental. While modern deep learning (DL) models can process longitudinal sequences end-to-end, their widespread clinical deployment is often hindered by opacity, high data requirements, and heavy computational footprints \cite{couplandExploringPotentialLimitations2025}. Consequently, explicit feature engineering remains critical. It is strictly necessary for interpretable, efficient tree-based models favored in clinical settings \cite{aroraImprovingClinicalDecision2026,zhangMedFeatModelAwareExplainabilityDriven2026}. Furthermore, explicitly engineered features can be injected into advanced DL architectures as static covariates, complementing learned representations to improve overall predictive performance and transparency \cite{limTimeseriesForecastingDeep2021,luSurveyDeepLearning2025}.

Historically, explicit feature engineering relied on costly \textit{manual} collaboration with healthcare professionals, or \textit{transformation-based} methods \cite{zhangOpenFEAutomatedFeature2023,hornAutofeatPythonLibrary2020,zhangMedFeatModelAwareExplainabilityDriven2026} that automate generation but lack clinical context and fundamentally struggle with irregular observation intervals. Recently, Large Language Models (LLMs) have emerged to bridge this gap. By leveraging vast domain knowledge, LLMs inject zero-shot clinical context, improving sample efficiency and generating transparent, rule-based features \cite{hollmannLargeLanguageModels2023,namOptimizedFeatureGeneration2024,abhyankarLLMFEAutomatedFeature2025}. However, existing LLM-based frameworks focus primarily on static tabular data and assume clean, structured inputs. Applying them to longitudinal EHRs, with their complex temporal dependencies, context-exceeding sequence lengths, and structural sparsity, remains a significant, unaddressed challenge.

To address these limitations, we introduce \method{}, a novel framework leveraging the clinical knowledge of LLMs to generate clinically motivated tabular features from irregularly sampled EHR time series. Crucially, \method{} addresses structural sparsity and variable frequencies by equipping the LLM with specialized programmatic tools to query and aggregate irregular temporal data. To reduce privacy exposure, the framework operates at the metadata level, exposing only the dataset schema and clinical task description to the LLM rather than raw patient-level records. The LLM generates executable feature-extraction code without direct access to sensitive patient records. During execution, \method{} transforms longitudinal measurements into engineered features that seamlessly integrate into standard ML pipelines. As illustrated in \autoref{fig:framework}, our framework executes both univariate and multivariate engineering to capture complex physiological interactions across multiple variables. The major contributions of this work are as follows:

\begin{itemize}
\item We propose \method{}, an LLM-empowered feature generation framework designed for irregularly sampled EHR time series. By exposing only dataset schemas and clinical task descriptions rather than raw records, our approach reduces privacy exposure to the LLM.
\item We directly address irregular sampling via a tool-augmented generation mechanism. By equipping the LLM with specialized temporal data-reading tools, it writes feature-extraction code that explicitly handles structural sparsity and variable measurement frequencies.
\item We develop a multi-source feature generation mechanism integrating LLM knowledge, task context, and inter-feature interactions through univariate and multivariate engineering. This is paired with an iterative strategy and in-loop validation, allowing the LLM to adaptively refine the generated features.
\item We implement a robust feature-to-code translation pipeline with syntax and runtime verification. We evaluate \method{} on eight clinical tasks across four datasets, achieving the highest mean AUROC on 7 out of 8 tasks (improving up to 6\% over baselines).
\end{itemize}

\section{Related Works}

In this section, we review feature engineering methods for tabular and time series data, with emphasis on approaches that generate explicit, inspectable features (including LLM-assisted pipelines). We intentionally exclude end-to-end learning-based representation methods (e.g., deep sequence encoders) because they learn latent embeddings rather than explicit engineered features, and therefore belong to a different methodological paradigm and evaluation setting than the one targeted in this work.

\paragraph{Transformation-Based Feature Engineering}
Transformation-based methods automate feature creation through predefined mathematical operations. For tabular data, frameworks like OpenFE \cite{zhangOpenFEAutomatedFeature2023}, AutoFeat \cite{hornAutofeatPythonLibrary2020}, and FETCH \cite{liLearningDataDrivenPolicy2022} successfully generate massive candidate spaces, relying on selection algorithms to filter for predictive utility. For time series, libraries extract temporal descriptors across statistical, spectral, and fractal domains (e.g., tsfresh \cite{christTimeSeriesFeatuRe2018}, TSFEL \cite{doncktTsflexFlexibleTime2021}, Kats \cite{Jiang_KATS_2022}, and Catch22 \cite{lubbaCatch22CAnonicalTimeseries2019}).

While effective at bulk generation, these methods suffer from two major limitations. First, they lack semantic awareness; they blindly apply mathematical transformations without understanding the domain context of the variables, often leading to a ``curse of dimensionality'' filled with redundant or clinically meaningless features. Second, they are overwhelmingly designed for regularly sampled signals. They fundamentally struggle with the highly irregular sampling rates, variable sequence lengths, and informative missingness that characterize clinical EHR time series, requiring ad-hoc imputation steps that can destroy underlying temporal patterns.

\paragraph{LLM-Based Feature Engineering}
Recent work has leveraged Large Language Models (LLMs) to inject semantic domain knowledge into feature engineering. Current approaches generally formulate this as a code-generation or rule-synthesis task. Methods like CAAFE \cite{hollmannLargeLanguageModels2023} and FeatLLM \cite{hanLargeLanguageModels2024} generate human-readable transformations from dataset descriptions or few-shot examples, while others like LLM-FE \cite{abhyankarLLMFEAutomatedFeature2025}, FeRG-LLM \cite{koFeRGLLMFeatureEngineering2025}, OCTree \cite{namOptimizedFeatureGeneration2024}, MedFeat \cite{zhangMedFeatModelAwareExplainabilityDriven2026}, and FAMOSE \cite{burghardtFAMOSEReActApproach2026} integrate iterative feedback mechanisms such as validation-driven refinement, evolutionary search, memory-guided selection, or ReAct-style reasoning-execution cycles to improve generated features \cite{yaoReActSynergizingReasoning2023}.

Despite these advancements, synthesizing the current literature reveals that LLM-based feature engineering remains constrained to static, tabular data. Adapting these pipelines to time series, particularly EHR data, presents unresolved challenges. First, existing methods operate on row-wise schema descriptions and lack the temporal reasoning required to generate complex longitudinal aggregations (e.g., tracking a patient's rolling health deterioration). Second, context window limits prevent LLMs from directly ingesting long, raw sequences of numerical time-series observations to derive patterns. Finally, existing LLM-generated code pipelines assume clean, structured inputs and are currently unequipped to autonomously write robust logic for the irregular intervals and missing values inherent to clinical data. Bridging this gap remains an open research challenge.

\section{FeatEHR-LLM}

\begin{figure}[t]
  \centering
  \includegraphics[width=0.95\textwidth]{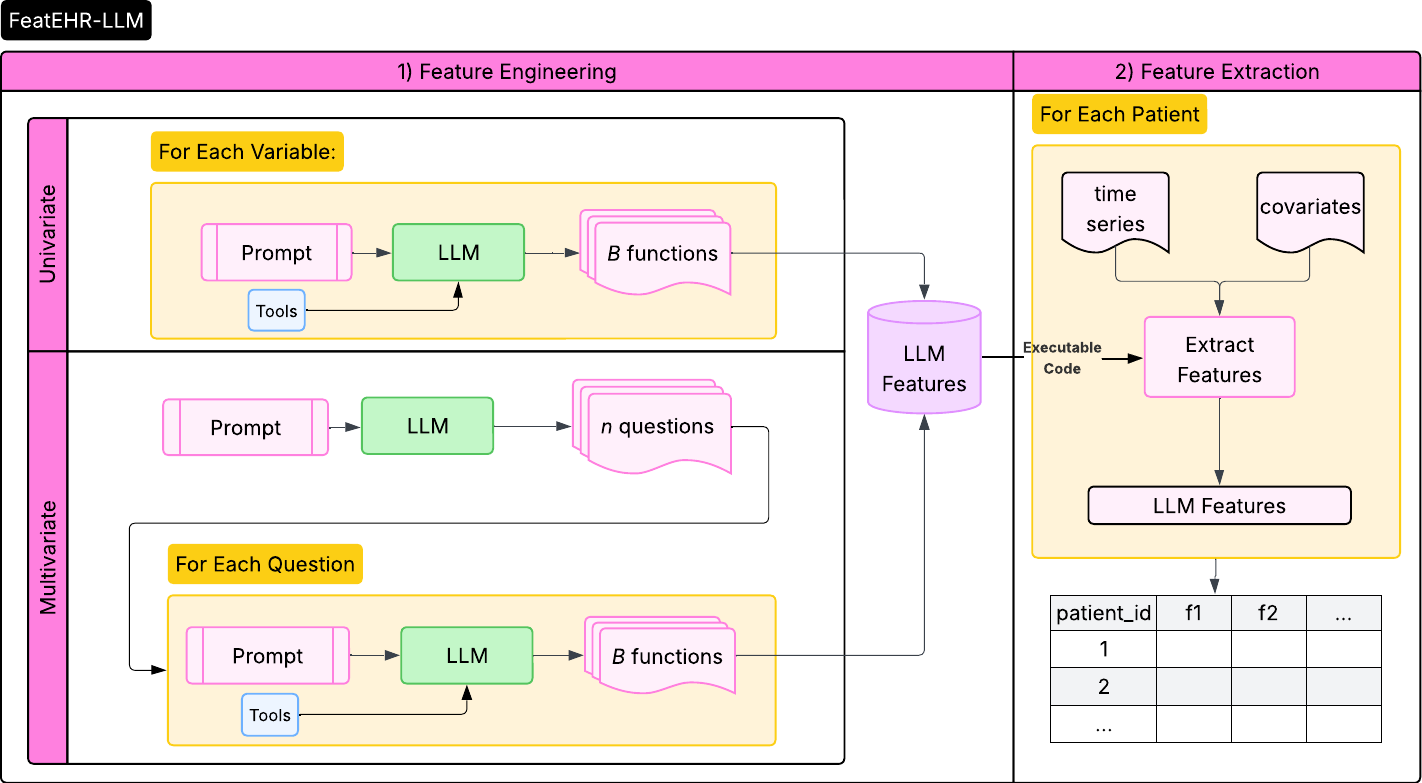}
  \caption{Overview of \method{}.}
  \label{fig:framework}
\end{figure}

\subsection{Problem Formulation}

We consider a supervised EHR prediction dataset $\mathcal{D} = \{(x_i, y_i)\}_{i=1}^{N}$, where $x_i = (c_i, T_i)$ denotes the input for patient $i$ and $y_i$ is the target outcome (e.g., in-hospital mortality). Here, $c_i$ collects static covariates such as demographics, while $T_i$ is an irregular multivariate time series containing physiological measurements and laboratory tests.

We represent the temporal record of patient $i$ as an ordered sequence
\[
T_i = \left(\left(t_{ik}, \{(v, x_{ikv})\}_{v \in \mathcal{O}_{ik}}\right)\right)_{k=1}^{m_i},
\]
where $t_{ik} \in \mathbb{R}_{\geq 0}$ is the $k$-th observation time, $\mathcal{O}_{ik}$ is the set of variables observed at that time, and $x_{ikv}$ is the value of variable $v$. This notation makes the irregularity explicit: both the number of observation times $m_i$ and the set of observed variables $\mathcal{O}_{ik}$ can vary across patients and across timestamps.

Let $\mathcal{X}$ denote the space of patient records $x_i=(c_i,T_i)$. For any subset of variables $S$, let $T_i^{(S)}$ denote the restriction of $T_i$ to measurements from variables in $S$.

Our goal is to learn a feature map $\phi: \mathcal{X} \rightarrow \mathbb{R}^d$ that converts each patient record into a fixed-length representation $z_i = \phi(x_i)$ suitable for a downstream predictor. Let $\mathcal{A}$ denote the model-training procedure (LightGBM with hyperparameter search), and define $h_{\phi}=\mathcal{A}(\phi(\mathcal{D}_{\mathrm{train}}))$. Given a validation split $\mathcal{D}_{\mathrm{val}}$, we compare feature maps through a task metric $\mathcal{E}$ (AUROC). We say that $\phi_1$ is better than $\phi_2$ if
\[
\mathcal{E}\big(h_{\phi_1}, \phi_1(\mathcal{D}_{\mathrm{val}})\big)
>
\mathcal{E}\big(h_{\phi_2}, \phi_2(\mathcal{D}_{\mathrm{val}})\big),
\]
where each $h_{\phi}$ is trained on the same split $\mathcal{D}_{\mathrm{train}}$.

Our framework also supports combining multiple feature maps. Given $K$ transformations $\phi_1, \dots, \phi_K$, we define the concatenated representation
\[
\phi_{\mathrm{cat}}(x) = \left[\phi_1(x); \phi_2(x); \dots; \phi_K(x)\right],
\]
where $[\cdot ; \cdot]$ denotes feature concatenation. This allows univariate and multivariate feature generators to contribute complementary information to the final predictor.

\subsection{Univariate Feature Engineering}

As illustrated in \autoref{fig:framework}, the univariate stage generates candidate features independently for each clinical variable. For a variable $v_i$ (e.g., heart rate or creatinine), we prompt the LLM with the task description together with lightweight metadata, including the variable name, unit, and summary statistics. The model then proposes $B$ executable Python functions, each of which takes the time series of a single variable for one patient and returns a scalar feature.

These functions are intended to be patient-agnostic: once generated, the same function is applied to every patient record. For each univariate function $f \in \mathcal{F}_{\mathrm{uni}}$, we denote by $v_f$ the variable used to generate it, and evaluate it as $f(T_i^{(\{v_f\})})$. We filter out invalid code by discarding functions that fail syntactic checks, and we retain the remaining candidates in a set $\mathcal{F}_{\mathrm{uni}}$. The prompt may additionally include helper routines for querying irregular time series, which encourages the LLM to generate functions that are compatible with sparse and unevenly sampled measurements. Algorithm~\ref{alg:univariate} summarizes the procedure, and concrete prompt and examples are provided in \autoref{fig:app-uni-prompt}.

\begin{algorithm}[H]
\caption{LLM-Based Univariate Feature Generation}
\begin{algorithmic}[1]
\REQUIRE Variables $\{v_1, \dots, v_m\}$, task description $\mathcal{T}$, LLM model, auxiliary tools, samples per prompt $B$
\ENSURE Set of valid generated feature functions $\mathcal{F}_{\text{uni}}$

\STATE Initialize $\mathcal{F}_{\text{uni}} \leftarrow \emptyset$

\FOR{each variable $v_i$ in $\{v_1, \dots, v_m\}$}
  \STATE Extract metadata tuple $(\texttt{name}_i, \texttt{unit}_i, \texttt{stats}_i)$
  \STATE Construct prompt $p_i \leftarrow \texttt{CreatePrompt}(\mathcal{T}, (\texttt{name}_i, \texttt{unit}_i, \texttt{stats}_i), \texttt{tools})$
  \STATE Generate $B$ candidate functions: $\{f_1, \dots, f_B\} \leftarrow \texttt{LLM}(p_i)$
  \FOR{each $f_j$ in $\{f_1, \dots, f_B\}$}
    \IF{$f_j$ is syntactically valid}
      \STATE $\mathcal{F}_{\text{uni}} \leftarrow \mathcal{F}_{\text{uni}} \cup \{f_j\}$
    \ENDIF
  \ENDFOR
\ENDFOR

\RETURN $\mathcal{F}_{\text{uni}}$
\end{algorithmic}
\label{alg:univariate}
\end{algorithm}

\subsection{Multivariate Feature Engineering}

The multivariate stage targets interactions that cannot be captured by a single variable alone. Instead of prompting directly for code, we first ask the LLM to propose $n_q$ clinically motivated questions relevant to the prediction task, together with the subset of variables needed to answer each question. This step uses the model's clinical prior knowledge to identify plausible cross-variable relationships before any feature code is generated.

For each question-variable pair $(q_j, V_j)$, we build a prompt containing the task description, the clinical question, and metadata for the variables in $V_j$. The LLM then produces $B$ candidate Python functions that compute scalar features from the corresponding subset of the patient record. As in the univariate stage, invalid programs are discarded. The resulting collection forms the multivariate feature set $\mathcal{F}_{\mathrm{multi}}$. For each multivariate function $g \in \mathcal{F}_{\mathrm{multi}}$, we denote by $V_g$ the variable subset associated with its originating question and evaluate it as $g(T_i^{(V_g)})$. Algorithm~\ref{alg:multivariate} provides a high-level view of this pipeline, and prompts and additional examples are given in \autoref{fig:app-multi-question} and \autoref{fig:app-multi-example}.

\begin{algorithm}[H]
\caption{LLM-Based Multivariate Feature Generation}
\begin{algorithmic}[1]
\REQUIRE Variables $\{v_1, \dots, v_m\}$, task description $\mathcal{T}$, LLM model, auxiliary tools, number of questions $n_q$, samples per prompt $B$
\ENSURE Set of valid multivariate feature functions $\mathcal{F}_{\text{multi}}$

\STATE Initialize $\mathcal{F}_{\text{multi}} \leftarrow \emptyset$

\STATE Generate clinically relevant question-variable pairs: $Q = \{(q_j, V_j)\}_{j=1}^{n_q} \leftarrow \texttt{LLMGenerateQuestions}(\mathcal{T}, \{(\texttt{name}_i, \texttt{unit}_i, \texttt{stats}_i)\}_{i=1}^m, n_q)$

\FOR{each pair $(q_j, V_j) \in Q$}
    \STATE Extract metadata tuple $(\texttt{name}_k, \texttt{unit}_k, \texttt{stats}_k)$ for each $v_k \in V_j$
    \STATE Construct prompt: $p_j \leftarrow \texttt{CreateMultivariatePrompt}(\mathcal{T}, q_j, V_j, \texttt{tools})$
    \STATE Generate $B$ candidate functions: $\{f_1, \dots, f_B\} \leftarrow \texttt{LLM}(p_j)$
    \FOR{each function $f_l \in \{f_1, \dots, f_B\}$}
    \IF{$f_l$ is syntactically valid and returns a scalar on valid input}
      \STATE $\mathcal{F}_{\text{multi}} \leftarrow \mathcal{F}_{\text{multi}} \cup \{f_l\}$
        \ENDIF
    \ENDFOR
\ENDFOR

\RETURN $\mathcal{F}_{\text{multi}}$
\end{algorithmic}
\label{alg:multivariate}
\end{algorithm}

\subsection{Feature Extraction}
We run Algorithms~\ref{alg:univariate} and \ref{alg:multivariate} for multiple rounds ($n_r$) to increase the diversity of the candidate function set. After generation, every valid function is executed on every patient record. For patient $i$, this yields the engineered representation

\[
z_i = \phi(x_i) = \left[\, f\big(T_i^{(\{v_f\})}\big) \;\middle|\; f \in \mathcal{F}_{\text{uni}} \,\right]
\oplus
\left[\, g\big(T_i^{(V_g)}\big) \;\middle|\; g \in \mathcal{F}_{\text{multi}} \,\right],
\]

where $\oplus$ denotes vector concatenation.

Stacking these representations across the cohort gives the design matrix

\[
Z = \left[ z_1^\top; z_2^\top; \dots; z_N^\top \right] \in \mathbb{R}^{N \times d},
\]

which is then used to train a downstream predictor such as LightGBM. In this way, the LLM is responsible for proposing executable feature maps, while model fitting remains entirely within a standard supervised learning pipeline.

\section{Experiments}
\subsection{Dataset}

\begin{table}[h]
  \small
  \centering
  \caption{Datasets details.}
  \import{.}{table_datasets.tex}
  \label{tab:datasets}
\end{table}

We leverage four publicly available ICU datasets encompassing eight prediction tasks. The PhysioNet 2012 Mortality Challenge (P12) comprises approximately 12,000 ICU admissions with 35 routinely measured time-series variables recorded during the first 48\,h, labeled for in-hospital mortality~\cite{silvaPredictingInHospitalMortalitya}. The PhysioNet 2019 Sepsis Challenge (P19) focuses on early sepsis detection and includes 40,336 patient stays with 34 physiological time-dependent variables~\cite{reynaEarlyPredictionSepsis2020}. For MIMIC-III, we adopt the well-established preprocessing pipeline from mimic3-benchmarks~\cite{wangMIMICExtractDataExtraction2020} to extract the first 48\,h of ICU data (41 variables) and predict in-hospital mortality (29,000 stays)~\cite{johnsonMIMICIIIFreelyAccessible2016}. From eICU-CRD \cite{pollardEICUCollaborativeResearch2018}, we use the FIDDLE preprocessing pipeline \cite{tangDemocratizingEHRAnalyses2020} to derive five tasks: in-hospital mortality at 48\,h and both acute respiratory failure (ARF) and shock prediction at 12\,h and 48\,h time horizons, covering between 110k and 145k ICU stays each with 95 variables. Table~\ref{tab:datasets} summarizes the dataset sizes, number of time-series variables, and positive class prevalence.

\subsection{Baselines}

We evaluated FeatEHR-LLM against several state-of-the-art feature engineering approaches:

\begin{itemize}
    \item \textbf{Baseline (BL):} This baseline extracts four statistical features (mean, standard deviation, minimum, and maximum) from each time series variable. Such statistical summarization is commonly used in many EHR studies and serves as a simple yet effective reference point for comparison.

    \item \textbf{tsfresh (TF):} This baseline leverages the \texttt{tsfresh} library \cite{christTimeSeriesFeatuRe2018}, which automatically extracts a comprehensive set of time series features, including statistical measures, autocorrelations, and Fourier coefficients. As a well-established method for time series feature extraction, it provides a strong baseline.

    \item \textbf{OpenFE:} OpenFE is a state-of-the-art feature engineering method for tabular data \cite{zhangOpenFEAutomatedFeature2023}. To adapt it for time series, it is applied to the BL features, which are first transformed into tabular format.

    \item \textbf{CAAFE and FeatLLM:} These are recent LLM-based feature engineering methods. CAAFE \cite{hollmannLargeLanguageModels2023} employs a context-aware pipeline to generate semantically meaningful features in the form of Python code, while FeatLLM \cite{hanLargeLanguageModels2024} utilizes few-shot prompting to create human-readable binary feature rules from a small set of examples. Both methods are adapted for time series data by applying them to the BL features.

    \item \textbf{Zeroshot:} This baseline uses the LLM in a zero-shot setting, where the LLM is asked to assign a risk score based on the task description and the time series data. Test size is set to 1000 samples to reduce inference costs in zeroshot setting.
\end{itemize}

We used LightGBM \cite{keLightGBMHighlyEfficient2017} as the prediction model for all downstream tasks, as it is a widely used and effective model for tabular data. For each task, we use a fixed split with 20\% held out as test data. The model is trained on the features generated by each method, and the performance is evaluated using Area Under the Receiver Operating Characteristic Curve (AUROC) as the primary metric. Confidence intervals were estimated via stratified bootstrap ($n_{boot}=1000$), sampling with replacement within each class to preserve class balance. Metrics were computed on each resample, and 95\% confidence intervals were obtained from the empirical 2.5th and 97.5th percentiles.

\subsection{Training Details}

We use Gemini-2.0-Flash \cite{Gemini25Flash} as the backbone LLM, which offers a reasonable trade-off between performance and computational cost. The temperature is set to 1.0 to balance creativity and consistency. For function generation, we sample $B=5$ candidate features per prompt and discard candidates with syntax errors. In multivariate feature generation, the number of questions is set to $n_q=20$. We further repeat both the univariate and multivariate feature generation procedures for $n_r=5$ rounds to increase the diversity of the generated features. For the downstream LightGBM classifier, we use grid search to tune the learning rate, number of estimators, maximum depth, and number of leaves.

\section{Results}

\subsection{Performance Gains}

\begin{table}[h]
  \small
  \centering
  \captionsetup{font=footnotesize}
  \caption{AUROC performance comparison across datasets and tasks. The best results are shown in bold, and the second-best are underlined. Subscript values indicate half-widths of the 95\% bootstrap confidence intervals, i.e., $(\mathrm{upper}-\mathrm{lower})/2$. (+) indicates that the features were added to the baseline feature set (BL).}
  \import{.}{table_main-baselines.tex}
  \label{tab:performance}
\end{table}

Table \ref{tab:performance} reports AUROC scores across eight benchmark tasks, comparing FeatEHR-LLM against strong baselines including OpenFE, FeatLLM, and CAAFE, all evaluated with the same base feature set (BL). FeatEHR-LLM achieves the highest mean AUROC on 7 out of 8 tasks. The gains are particularly pronounced in the P19 and MIMIC-III datasets, with improvements of 6 and 3.1 percentage points over the baseline, respectively. A similar trend appears across MIMIC-III, eICU, and PhysioNet tasks. Overall, FeatEHR-LLM is effective at synthesizing informative features from EHRs for clinical prediction tasks.

\paragraph{Compare with zeroshot}
Table \ref{tab:zeroshot} shows the performance of our method, FeatEHR-LLM, compared to the zero-shot LLMs on three datasets: p12s, p19, and MIMIC-III. The results are presented in terms of Area Under the Receiver Operating Characteristic Curve (AUC) and F1 score. The numbers in parentheses indicate the standard deviation across five runs.

\begin{table}[h]
  \small
  \centering
  \caption{Comparison with zeroshot across 3 datasets. Test size is 1000 samples.}
  \import{.}{table_zeroshot.tex}
  \label{tab:zeroshot}
\end{table}

\subsection{Generalizability}
We evaluate the generalizability of \method{} across three dimensions: dataset size, backbone language model, and downstream prediction architecture.

\begin{figure}[ht]
  \centering
  \includegraphics[width=0.75\textwidth]{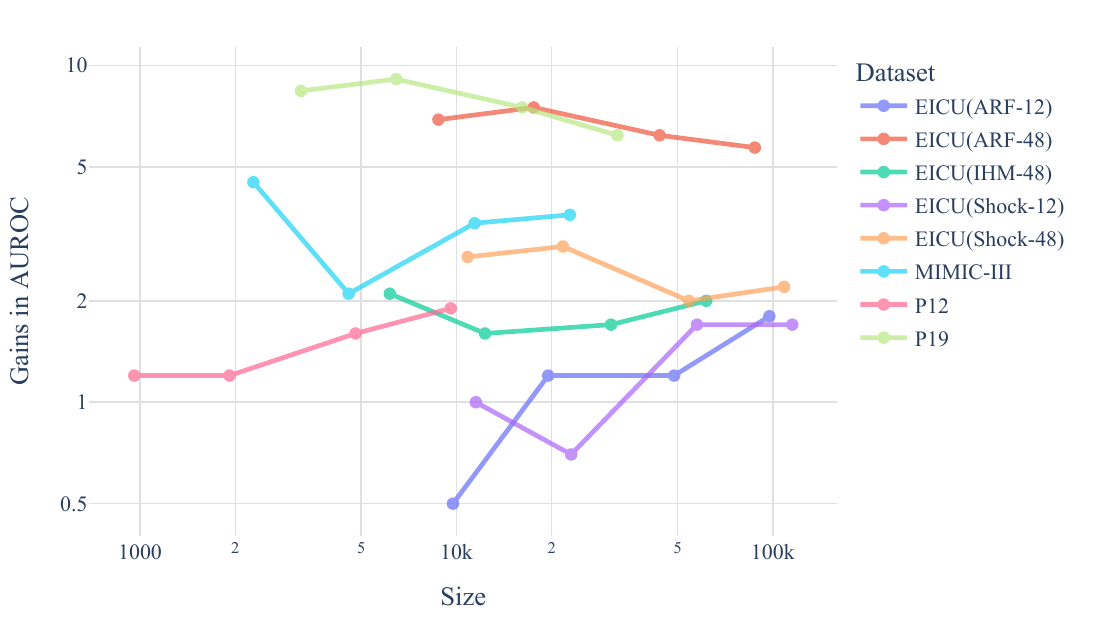}
  \caption{Performance gain over baselines across different dataset sizes. The x-axis represents the dataset size, while the y-axis shows the performance gain in terms of AUROC.}
  \label{fig:gen-size}
\end{figure}

\paragraph{Dataset Size}
Figure~\ref{fig:gen-size} shows the AUROC improvement from augmenting the baseline features (BL) with \method{} across varying dataset sizes (10\%, 20\%, 50\%, and 100\%). The results indicate that \method{} yields consistent performance improvements across all regimes, including under low-resource conditions. Notably, even with only 10\% of the data, the feature augmentation provides a meaningful lift over the baseline, suggesting strong data efficiency and generalization under limited supervision.

\begin{figure}[ht]
  \centering
  \includegraphics[width=0.75\textwidth]{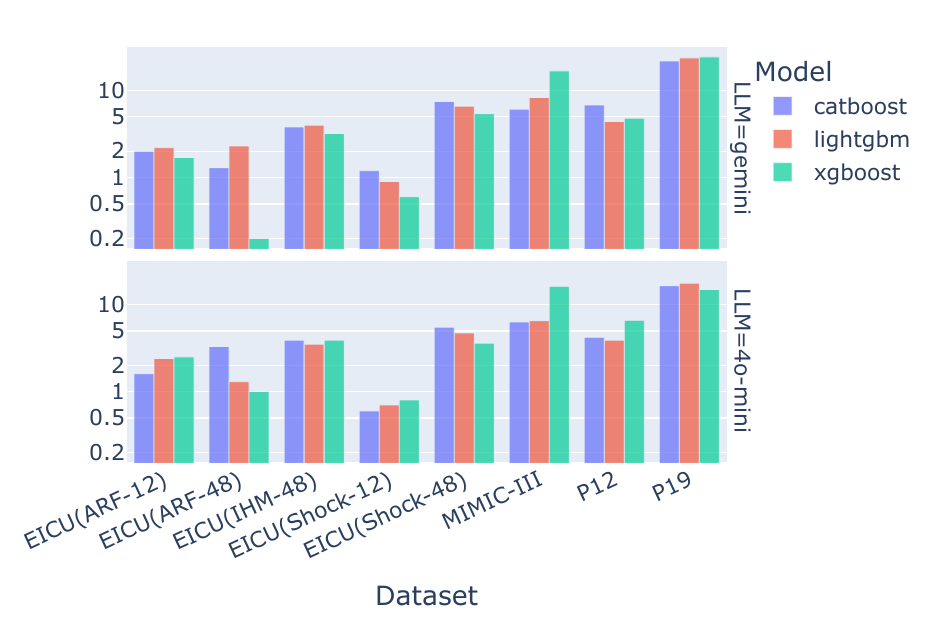}
  \caption{Performance gain over baselines across different dataset sizes. The x-axis represents the dataset size, while the y-axis shows the performance gain in terms of AUROC.}
  \label{fig:gen-model}
\end{figure}

\paragraph{Choice of LLM and Downstream Model}
Figure~\ref{fig:gen-model} reports the AUROC improvements obtained by adding \method{} to the baseline features (BL) across multiple downstream classifiers: LightGBM, XGBoost, and CatBoost, and across two LLMs: gemini and gpt-4o-mini. \method{} improves AUROC across the evaluated model settings, with the largest improvements observed for LightGBM. These results indicate compatibility of the generated features with diverse learning algorithms.

\paragraph{Ablation Study}
Table~\ref{tab:ablation} presents the results of our ablation study. Here, \textit{B=1} refers to generating only one feature per prompt, while \textit{B=5} means generating five features and using all of them. The \textit{best of B} variant selects the single feature with the highest mutual information with the target variable from the set of B generated features. We observe that generating multiple features (\textit{B=5}) generally improves performance over the other two methods.

The variant \textit{single round} indicates that we run the feature generation process only once ($n_r=1$) compared to the full method (\textit{multiple rounds}, $n_r=5$). The results show that running multiple rounds of feature generation leads to better performance, suggesting that iterative refinement is beneficial for feature quality.

\method{} achieves the highest mean AUROC among the listed ablation variants in 7 out of 8 tasks (with a tie in one task). The results indicate that removing the univariate feature generation step (\textit{w/o uni}) leads to a larger performance drop than removing the multivariate step (\textit{w/o multi}), suggesting that univariate feature generation contributes more strongly to overall performance in this setup.

\begin{table}[h]
  \small
  \centering
  \caption{Ablation study. AUROC performance comparison across datasets and tasks. The best results are shown in bold, and the second-best are underlined. Subscript values indicate half-widths of the 95\% bootstrap confidence intervals, i.e., $(\mathrm{upper}-\mathrm{lower})/2$. A (+) indicates that features were added to the baseline feature set (BL).}
  \import{.}{table_main-ablations.tex}
  \label{tab:ablation}
\end{table}

\section{Discussion}

\paragraph{Scope of irregularity}
The irregularity addressed by \method{} is the observational irregularity typical of EHR time series: variables are measured at nonuniform times, different variables are observed at different timestamps, and the number of recorded observations varies across patients even within a fixed task-specific horizon \cite{liptonLearningDiagnoseLSTM2017,shuklaMultiTimeAttentionNetworks2021}. Our tool-augmented feature generation is designed to reason over these uneven observation patterns directly. However, the framework does not explicitly model statistical missingness mechanisms such as MCAR, MAR, or MNAR, nor does it infer why a clinician chose not to measure a variable \cite{littleStatisticalAnalysisMissing2019a,agnielBiasesElectronicHealth2018}. This distinction is important because, in our setting, absent measurements are treated primarily as structural sparsity in the observation process rather than as separately parameterized missing-data processes.

\paragraph{Redundant features}
Generating multiple candidate feature extraction scripts per variable or clinical question yields a set of related features that often differ only slightly in thresholds or aggregation logic. Although this introduces redundancy, our ablation results suggest that these variants are beneficial rather than purely superfluous. In tree-based predictors such as gradient boosted decision trees, correlated features can still contribute complementary signal because they induce different partitions of the feature space and enable alternative splitting rules during training. Consequently, multiple threshold variants may capture distinct regions of the data distribution and facilitate the discovery of informative interactions with other variables. This behavior is consistent with prior work showing that tree ensembles distribute predictive importance across groups of correlated variables and can flexibly select among them when constructing decision trees \cite{lundbergConsistentIndividualizedFeature2019}. More broadly, the machine learning literature has shown that diversity in feature representations can improve predictive performance by providing complementary views of the same underlying signal, an idea closely related to attribute bagging and representation ensembles \cite{bryllAttributeBaggingImproving2003}. Empirical studies of feature selection in tree ensembles further demonstrate that predictive importance is often spread across correlated feature groups, implying that removing such features too aggressively may eliminate useful alternative decision boundaries \cite{liuControlBurnFeatureSelection2021}. Together, these findings suggest that LLM-generated feature variants can act as complementary representations of clinical signals, increasing the representational flexibility available to the downstream model.

\paragraph{Computational cost}
Our framework increases computation during feature generation and extraction because it relies on multiple rounds and multiple candidate programs per prompt. The number of LLM calls scales with the number of variables, the number of multivariate questions, the number of rounds, and the number of samples per prompt. Nevertheless, compared with zero-shot or embedding-based approaches that require serializing each patient's time series into text, our method is substantially more efficient at inference time from the LLM perspective. The LLM only receives metadata and the task description, which keeps the prompt short and makes generation independent of the number of patients. The main cost is therefore concentrated in feature proposal rather than patient-level prompting.

\paragraph{Clinical interpretability}
A practical advantage of our approach is that the generated features are expressed as explicit, executable rules rather than opaque latent representations. This makes them easier to inspect than features produced by end-to-end representation learning methods \cite{frascaExplainableInterpretableArtificial2024}. However, interpretability should not be overstated: understandable code does not automatically imply clinical validity \cite{elshawiInterpretabilityMachineLearningbased2019}. Moreover, because the features are proposed by an LLM, generation can still be affected by hallucination, and this risk may increase at higher sampling temperatures that favor diversity and creativity \cite{zhouLargeLanguageModels2025, kimMedicalHallucinationsFoundation2025,roustanCliniciansGuideLarge2025}. In the clinical setting, such plausible but incorrect feature definitions are particularly concerning because they may encode unsafe assumptions while remaining superficially interpretable \cite{pilgramMagnitudeImpactHallucinations2025}. For this reason, generated features should be validated empirically and reviewed by clinicians or domain experts before use, especially when they may influence high-stakes decisions \cite{zhouLargeLanguageModels2025}.

\section{Conclusion}

We presented \method{}, a framework that leverages LLMs to engineer clinically motivated tabular features from irregularly sampled EHR time series. By operating exclusively on dataset metadata rather than raw records, the framework limits patient privacy exposure while producing executable feature-extraction code through a tool-augmented mechanism designed to handle structural sparsity and variable observation frequencies. Multi-round univariate and multivariate generation, coupled with in-loop syntax and runtime validation, yields diverse and robust feature sets. Across eight clinical prediction tasks on four ICU datasets, \method{} achieves the highest mean AUROC on 7 out of 8 tasks, with gains of up to 6 percentage points over baselines. Generated features remain fully interpretable as executable code and integrate seamlessly into standard ML pipelines. Key limitations include the risk of LLM hallucinations in generated feature logic and the absence of explicit missing-data mechanism modeling, both of which warrant expert review before clinical deployment. Future work could extend the framework to additional EHR modalities and incorporate automated clinical validation feedback.

\section*{Acknowledgments}
This work was supported by the RealCare project, which has received funding from the European Union and by the Swiss State Secretariat for Education, Research and Innovation (SERI). Additionally, we would like to thank the EPFL IC Cluster for providing the computational resources used in this work.

\appendix

\section{Prompts and Examples}

\begin{figure}[h]
  \centering
  \captionsetup{font=small}
  \includegraphics[width=1.0\textwidth]{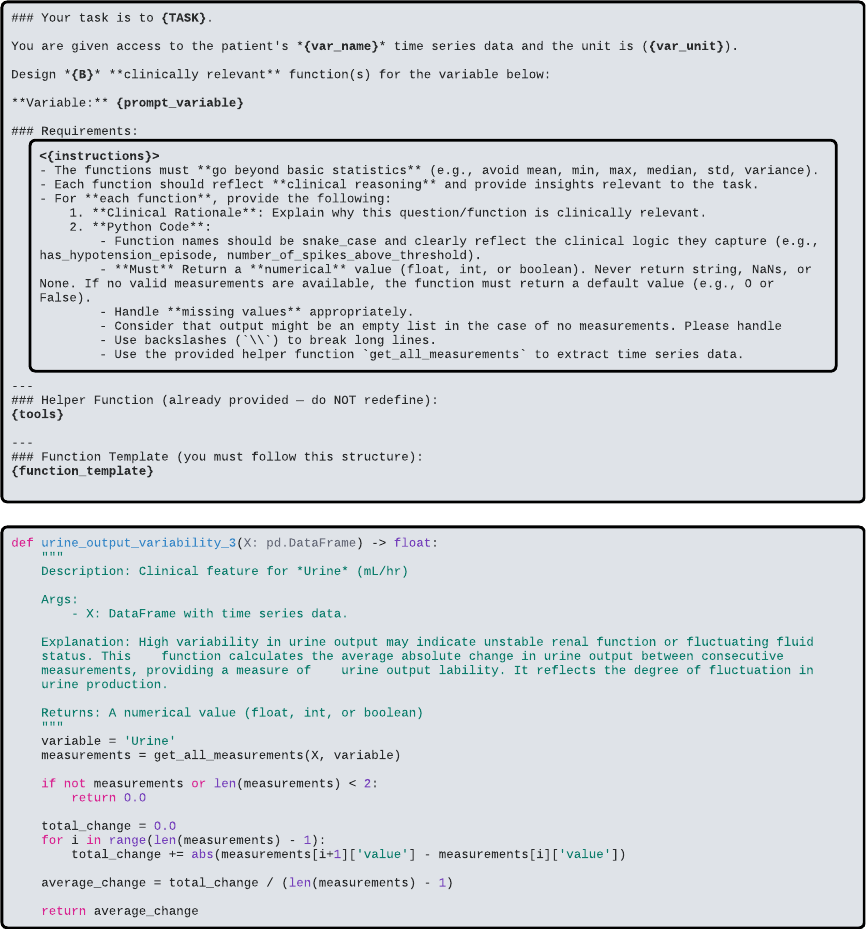}
  \caption{Univariate feature engineering step. Top: prompt used to generate candidate univariate feature functions. Bottom: example of a generated function for urine.}
  \label{fig:app-uni-prompt}
\end{figure}

\begin{figure}[h]
  \centering
  \captionsetup{font=small}
  \includegraphics[width=1.0\textwidth]{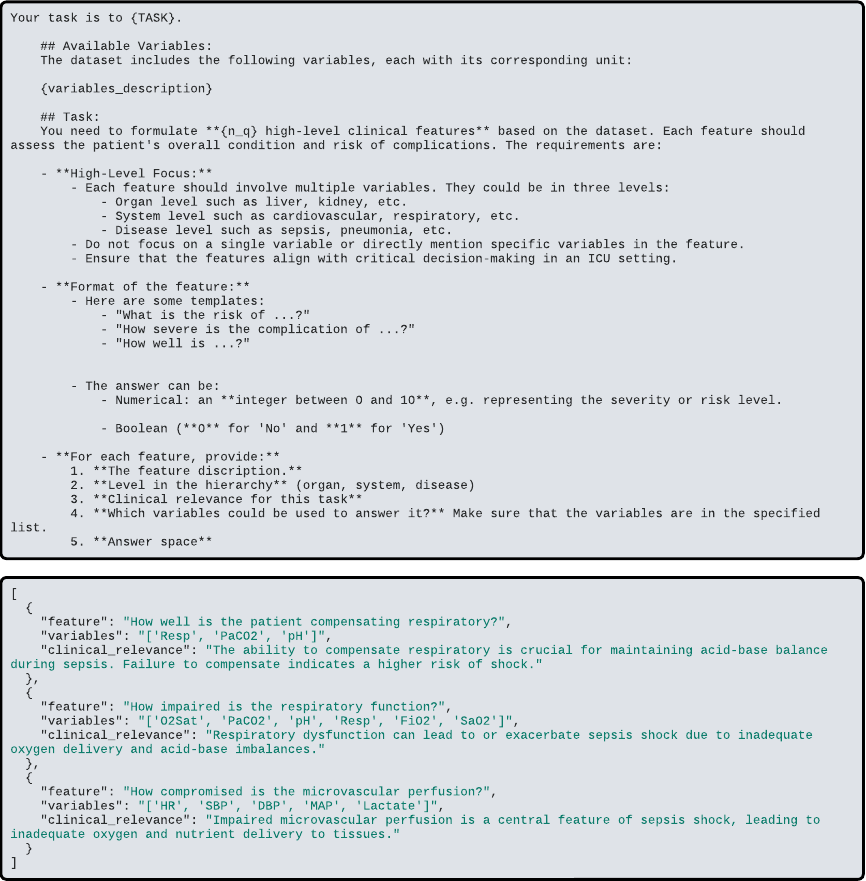}
  \caption{Multivariate feature engineering step. Top: prompt used to generate clinically relevant questions. Bottom: three example questions for the P12 dataset. Each question is paired with the subset of variables that the LLM deems necessary to answer it, along with its clinical relevance to the specified task.}
  \label{fig:app-multi-question}
\end{figure}

\begin{figure}[h]
  \centering
  \captionsetup{font=small}
  \includegraphics[width=1.0\textwidth]{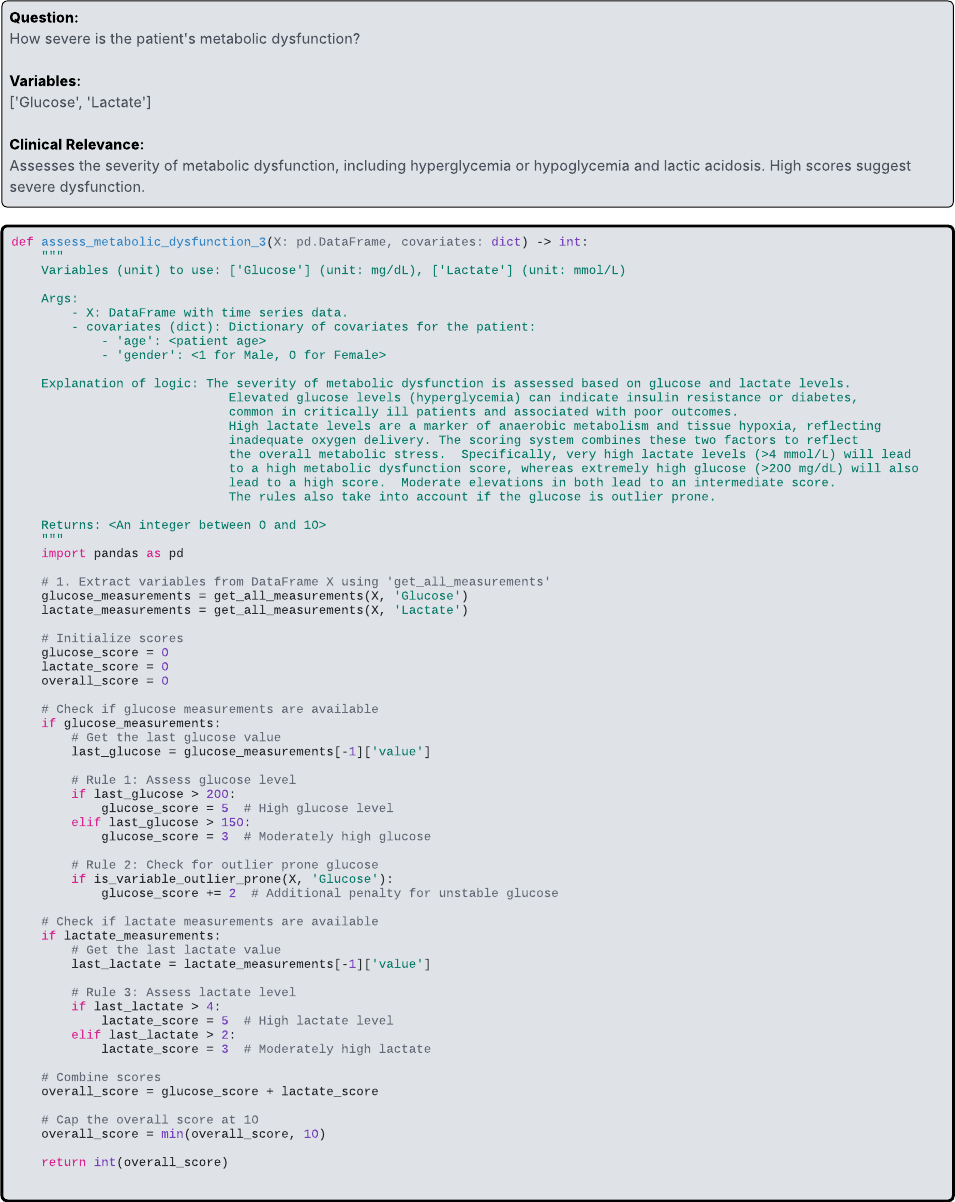}
  \caption{Multivariate feature engineering example. Top: generated question and required variables. Bottom: example of a generated feature function that answers the question. The function explains its logic in the docstring: it first reads the measurements using the specified tool and then applies the corresponding decision logic.}
  \label{fig:app-multi-example}
\end{figure}

\begin{figure}[h]
  \centering
  \captionsetup{font=small}
  \includegraphics[width=1.0\textwidth]{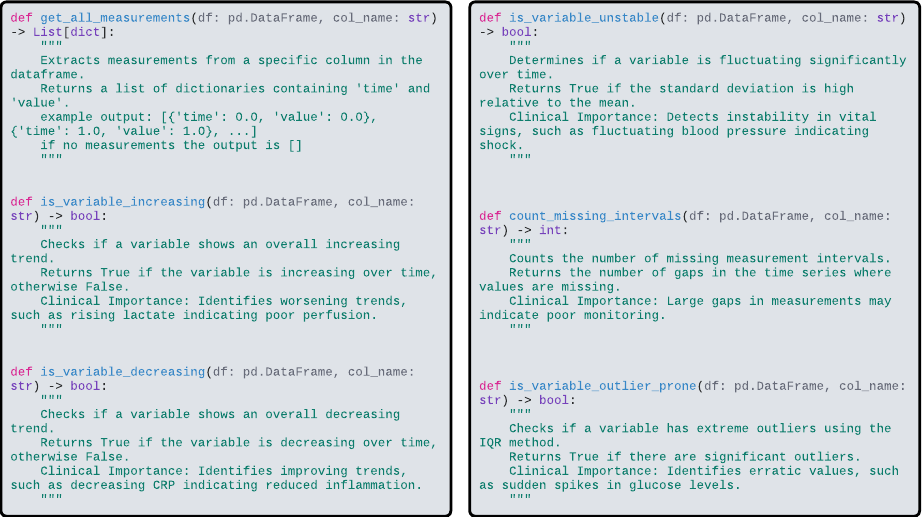}
  \caption{Tool functions available to the LLM. Univariate feature engineering uses only \texttt{get\_all\_measurements}, whereas multivariate feature engineering uses the full tool set.}
  \label{fig:app-tools}
\end{figure}

\bibliographystyle{unsrtnat}
\bibliography{zotero2}

\end{document}

%% file: table_datasets.tex
\begin{tabular}{lccc}
\toprule
Dataset & Size &  \texttt{n\_ts} & Prevalence(\%) \\
\midrule
P12 & 12k & 35 & 12.89 \\
P19 & 40k & 34 & 7.1 \\
MIMIC-III(IHM-48) & 29k & 41 & 8.87 \\
EICU(IHM-48) & 77k & 95 & 9.13 \\
EICU(ARF-12) & 123k & 95 & 4.5 \\
EICU(Shock-12) & 145k & 95 & 3.93 \\
EICU(ARF-48) & 110k & 95 & 1.81 \\
EICU(Shock-48) & 136k & 95 & 4.17 \\
\bottomrule
\end{tabular}

%% file: table_main-baselines.tex
\begin{tabular}{lcccccc}
\toprule
Dataset & +FeatEHR-LLM & +FeatLLM & +CAAFE & +TF & +OpenFE & BL \\
\midrule
P12 & \textbf{85.9}$_{1.9}$ & 84.6$_{2.0}$ & 84.3$_{2.0}$ & \underline{85.6}$_{2.0}$ & 85.0$_{1.9}$ & 83.6$_{2.0}$ \\
P19 & \textbf{92.1}$_{1.3}$ & 86.0$_{1.6}$ & 86.3$_{1.5}$ & \underline{90.1}$_{1.4}$ & 88.6$_{1.4}$ & 86.4$_{1.5}$ \\
MIMIC-III & \textbf{94.1}$_{1.0}$ & 90.4$_{1.4}$ & 90.4$_{1.4}$ & \underline{92.6}$_{1.2}$ & 91.2$_{1.3}$ & 91.0$_{1.3}$ \\
EICU(IHM-48) & \textbf{83.8}$_{0.9}$ & 82.1$_{1.0}$ & \underline{82.2}$_{1.0}$ & 82.0$_{1.0}$ & 81.7$_{1.0}$ & 81.8$_{0.9}$ \\
EICU(ARF-12) & \textbf{78.5}$_{1.2}$ & 76.9$_{1.2}$ & 77.0$_{1.3}$ & \underline{78.2}$_{1.2}$ & 76.2$_{1.4}$ & 76.9$_{1.3}$ \\
EICU(Shock-12) & \textbf{77.5}$_{1.2}$ & 76.2$_{1.2}$ & 76.1$_{1.2}$ & \underline{76.5}$_{1.2}$ & 75.9$_{1.2}$ & 76.4$_{1.2}$ \\
EICU(ARF-48) & \underline{84.1}$_{1.5}$ & 78.3$_{2.0}$ & 78.9$_{2.0}$ & \textbf{84.6}$_{1.6}$ & 77.1$_{2.0}$ & 78.0$_{2.1}$ \\
EICU(Shock-48) & \textbf{86.8}$_{1.0}$ & 84.5$_{1.0}$ & 84.6$_{1.0}$ & \underline{86.0}$_{0.9}$ & 84.4$_{1.1}$ & 84.7$_{1.0}$ \\
\bottomrule
\end{tabular}


%% file: table_zeroshot.tex
\begin{tabular}{lcccc}
\toprule
Dataset & +FeatEHR-LLM & BL & FeatEHR-LLM & Zeroshot \\
\midrule
P12 & \textbf{86.9}$_{1.9}$ & 84.8$_{2.1}$ & \underline{85.8}$_{2.0}$ & 73.1$_{4.1}$ \\
P19 & \underline{92.5}$_{1.2}$ & 85.5$_{1.6}$ & \textbf{92.7}$_{1.3}$ & 62.2$_{6.6}$ \\
MIMIC-III & \textbf{95.6}$_{0.7}$ & 93.3$_{1.0}$ & \underline{93.6}$_{1.0}$ & 88.2$_{3.2}$ \\
\bottomrule
\end{tabular}


%% file: table_main-ablations.tex


\begin{tabular}{lccc|c|cc}
\toprule
Dataset & FeatEHR-LLM & best of B & B=1 & single round  & w/o multi  & w/o uni  \\
& (B=5) & & & & & \\
\midrule
P12 & \textbf{85.7$_{1.8}$} & \underline{85.5$_{1.9}$} & 82.8$_{2.1}$ & 82.9$_{2.1}$ & 84.8$_{2.0}$ & 83.4$_{1.9}$ \\
P19 & \textbf{92.5$_{1.2}$} & 92.1$_{1.2}$ & 85.3$_{1.6}$ & 90.9$_{1.3}$ & \underline{92.4$_{1.2}$} & 83.8$_{1.5}$ \\
MIMIC-III & \textbf{92.9$_{1.1}$} & 92.2$_{1.2}$ & 89.8$_{1.5}$ & 91.8$_{1.2}$ & \underline{92.5$_{1.1}$} & 90.9$_{1.3}$ \\
EICU(IHM-48) & \textbf{83.4$_{0.9}$} & \underline{82.8$_{0.9}$} & 81.6$_{1.0}$ & 82.3$_{0.9}$ & 82.6$_{0.9}$ & 81.1$_{1.0}$ \\
EICU(ARF-12) & \textbf{74.6$_{1.3}$} & 74.0$_{1.3}$ & 68.6$_{1.4}$ & 72.0$_{1.3}$ & \underline{74.5$_{1.3}$} & 69.0$_{1.4}$ \\
EICU(Shock-12) & \textbf{74.7$_{1.4}$} & 73.1$_{1.3}$ & 70.6$_{1.3}$ & 73.0$_{1.3}$ & \underline{74.0$_{1.3}$} & 70.5$_{1.4}$ \\
EICU(ARF-48) & 82.5$_{1.7}$ & \textbf{82.9$_{1.7}$} & 73.1$_{2.2}$ & 81.5$_{1.6}$ & \underline{82.7$_{1.7}$} & 73.7$_{2.3}$ \\
EICU(Shock-48) & \textbf{85.6$_{1.0}$} & 84.1$_{1.1}$ & 81.8$_{1.2}$ & 82.8$_{1.1}$ & \textbf{85.6$_{1.0}$} & 80.1$_{1.3}$ \\
\bottomrule
\end{tabular}

%% file: zotero2.bib
@misc{abhyankarLLMFEAutomatedFeature2025,
  title = {{{LLM-FE}}: {{Automated Feature Engineering}} for {{Tabular Data}} with {{LLMs}} as {{Evolutionary Optimizers}}},
  shorttitle = {{{LLM-FE}}},
  author = {Abhyankar, Nikhil and Shojaee, Parshin and Reddy, Chandan K.},
  year = 2025,
  month = mar,
  number = {arXiv:2503.14434},
  eprint = {2503.14434},
  primaryclass = {cs},
  publisher = {arXiv},
  doi = {10.48550/arXiv.2503.14434},
  urldate = {2025-04-01},
  archiveprefix = {arXiv}
}

@article{agnielBiasesElectronicHealth2018,
  title = {Biases in Electronic Health Record Data Due to Processes within the Healthcare System: Retrospective Observational Study},
  shorttitle = {Biases in Electronic Health Record Data Due to Processes within the Healthcare System},
  author = {Agniel, Denis and Kohane, Isaac S. and Weber, Griffin M.},
  year = 2018,
  month = apr,
  journal = {BMJ},
  volume = {361},
  pages = {k1479},
  address = {Clinical research ed.},
  issn = {1756-1833},
  doi = {10.1136/bmj.k1479},
  langid = {english},
  pmcid = {PMC5925441},
  pmid = {29712648}
}

@article{aroraImprovingClinicalDecision2026,
  title = {Improving Clinical Decision Support through Interpretable Machine Learning and Error Handling in Electronic Health Records},
  author = {Arora, Mehak and Mortagy, Hassan and Dwarshuis, Nathan and Wang, Jeffrey and Yang, Philip and Holder, Andre L and Gupta, Swati and Kamaleswaran, Rishikesan},
  year = 2026,
  month = jan,
  journal = {Journal of the American Medical Informatics Association},
  volume = {33},
  number = {1},
  pages = {123--132},
  issn = {1527-974X},
  doi = {10.1093/jamia/ocaf058},
  urldate = {2026-03-12}
}

@article{bryllAttributeBaggingImproving2003,
  title = {Attribute Bagging: Improving Accuracy of Classifier Ensembles by Using Random Feature Subsets},
  shorttitle = {Attribute Bagging},
  author = {Bryll, Robert and {Gutierrez-Osuna}, Ricardo and Quek, Francis},
  year = 2003,
  month = jun,
  journal = {Pattern Recognition},
  volume = {36},
  number = {6},
  pages = {1291--1302},
  issn = {0031-3203},
  doi = {10.1016/S0031-3203(02)00121-8},
  urldate = {2026-03-12}
}

@misc{burghardtFAMOSEReActApproach2026,
  title = {{{FAMOSE}}: {{A ReAct Approach}} to {{Automated Feature Discovery}}},
  shorttitle = {{{FAMOSE}}},
  author = {Burghardt, Keith and Liu, Jienan and Sakib, Sadman and Hao, Yuning and Li, Bo},
  year = 2026,
  month = feb,
  number = {arXiv:2602.17641},
  eprint = {2602.17641},
  primaryclass = {cs},
  publisher = {arXiv},
  doi = {10.48550/arXiv.2602.17641},
  urldate = {2026-03-12},
  archiveprefix = {arXiv}
}

@article{chenHarnessingPowerClinical2023,
  title = {Harnessing the Power of Clinical Decision Support Systems: Challenges and Opportunities},
  shorttitle = {Harnessing the Power of Clinical Decision Support Systems},
  author = {Chen, Zhao and Liang, Ning and Zhang, Haili and Li, Huizhen and Yang, Yijiu and Zong, Xingyu and Chen, Yaxin and Wang, Yanping and Shi, Nannan},
  year = 2023,
  month = nov,
  journal = {Open Heart},
  volume = {10},
  number = {2},
  publisher = {British Cardiovascular Society},
  issn = {2053-3624},
  doi = {10.1136/openhrt-2023-002432},
  urldate = {2026-03-12},
  copyright = {This is an open access article distributed in accordance with the Creative Commons Attribution 4.0 Unported (CC BY 4.0) license},
  langid = {english},
  pmid = {10.1136/openhrt-2023-002432}
}

@article{christTimeSeriesFeatuRe2018,
  title = {Time {{Series FeatuRe Extraction}} on Basis of {{Scalable Hypothesis}} Tests (Tsfresh -- {{A Python}} Package)},
  shorttitle = {Tsfresh},
  author = {Christ, Maximilian and Braun, Nils and Neuffer, Julius and {Kempa-Liehr}, Andreas W.},
  year = 2018,
  month = sep,
  journal = {Neurocomputing},
  volume = {307},
  pages = {72--77},
  issn = {0925-2312},
  doi = {10.1016/j.neucom.2018.03.067},
  urldate = {2025-06-23}
}

@article{couplandExploringPotentialLimitations2025,
  title = {Exploring the Potential and Limitations of Deep Learning and Explainable {{AI}} for Longitudinal Life Course Analysis},
  author = {Coupland, Helen and Scheidwasser, Neil and Katsiferis, Alexandros and Davies, Megan and Flaxman, Seth and Hulvej Rod, Naja and Mishra, Swapnil and Bhatt, Samir and Unwin, H. Juliette T.},
  year = 2025,
  month = apr,
  journal = {BMC Public Health},
  volume = {25},
  number = {1},
  pages = {1520},
  issn = {1471-2458},
  doi = {10.1186/s12889-025-22705-4},
  urldate = {2026-03-12},
  langid = {english}
}

@misc{doncktTsflexFlexibleTime2021,
  title = {Tsflex: Flexible Time Series Processing \& Feature Extraction},
  shorttitle = {Tsflex},
  author = {Donckt, Jonas Van Der and Donckt, Jeroen Van Der and Deprost, Emiel and Hoecke, Sofie Van},
  year = 2021,
  month = dec,
  number = {arXiv:2111.12429},
  eprint = {2111.12429},
  primaryclass = {cs},
  publisher = {arXiv},
  doi = {10.48550/arXiv.2111.12429},
  urldate = {2025-06-23},
  archiveprefix = {arXiv}
}

@article{elshawiInterpretabilityMachineLearningbased2019,
  title = {On the Interpretability of Machine Learning-Based Model for Predicting Hypertension},
  author = {Elshawi, Radwa and {Al-Mallah}, Mouaz H. and Sakr, Sherif},
  year = 2019,
  month = jul,
  journal = {BMC Medical Informatics and Decision Making},
  volume = {19},
  number = {1},
  pages = {146},
  issn = {1472-6947},
  doi = {10.1186/s12911-019-0874-0},
  urldate = {2026-03-13},
  langid = {english}
}

@article{frascaExplainableInterpretableArtificial2024,
  title = {Explainable and Interpretable Artificial Intelligence in Medicine: A Systematic Bibliometric Review},
  shorttitle = {Explainable and Interpretable Artificial Intelligence in Medicine},
  author = {Frasca, Maria and La Torre, Davide and Pravettoni, Gabriella and Cutica, Ilaria},
  year = 2024,
  month = feb,
  journal = {Discover Artificial Intelligence},
  volume = {4},
  number = {1},
  pages = {15},
  issn = {2731-0809},
  doi = {10.1007/s44163-024-00114-7},
  urldate = {2026-03-13},
  langid = {english}
}

@misc{Gemini25Flash,
  title = {Gemini 2.5 {{Flash}}},
  journal = {Google DeepMind},
  urldate = {2025-06-24},
  howpublished = {https://deepmind.google/models/gemini/flash/},
  langid = {english}
}

@article{getzenMiningEquitableHealth2023a,
  title = {Mining for Equitable Health: {{Assessing}} the Impact of Missing Data in Electronic Health Records},
  shorttitle = {Mining for Equitable Health},
  author = {Getzen, Emily and Ungar, Lyle and Mowery, Danielle and Jiang, Xiaoqian and Long, Qi},
  year = 2023,
  month = mar,
  journal = {Journal of Biomedical Informatics},
  volume = {139},
  pages = {104269},
  issn = {1532-0464},
  doi = {10.1016/j.jbi.2022.104269},
  urldate = {2026-03-12}
}

@article{ghassemiReviewChallengesOpportunities2020,
  title = {A {{Review}} of {{Challenges}} and {{Opportunities}} in {{Machine Learning}} for {{Health}}},
  author = {Ghassemi, Marzyeh and Naumann, Tristan and Schulam, Peter and Beam, Andrew L. and Chen, Irene Y. and Ranganath, Rajesh},
  year = 2020,
  month = may,
  journal = {AMIA Summits on Translational Science Proceedings},
  volume = {2020},
  pages = {191--200},
  issn = {2153-4063},
  urldate = {2021-12-31},
  pmcid = {PMC7233077},
  pmid = {32477638}
}

@misc{hanLargeLanguageModels2024,
  title = {Large {{Language Models Can Automatically Engineer Features}} for {{Few-Shot Tabular Learning}}},
  shorttitle = {{{FeatLLM}}},
  author = {Han, Sungwon and Yoon, Jinsung and Arik, Sercan O. and Pfister, Tomas},
  year = 2024,
  month = may,
  number = {arXiv:2404.09491},
  eprint = {2404.09491},
  primaryclass = {cs},
  publisher = {arXiv},
  doi = {10.48550/arXiv.2404.09491},
  urldate = {2025-02-19},
  archiveprefix = {arXiv}
}

@misc{hollmannLargeLanguageModels2023,
  title = {Large {{Language Models}} for {{Automated Data Science}}: {{Introducing CAAFE}} for {{Context-Aware Automated Feature Engineering}}},
  shorttitle = {{{CAAFE}}},
  author = {Hollmann, Noah and M{\"u}ller, Samuel and Hutter, Frank},
  year = 2023,
  month = sep,
  number = {arXiv:2305.03403},
  eprint = {2305.03403},
  primaryclass = {cs},
  publisher = {arXiv},
  doi = {10.48550/arXiv.2305.03403},
  urldate = {2025-04-21},
  archiveprefix = {arXiv}
}

@misc{hornAutofeatPythonLibrary2020,
  title = {The Autofeat {{Python Library}} for {{Automated Feature Engineering}} and {{Selection}}},
  shorttitle = {{{AutoFeat}}},
  author = {Horn, Franziska and Pack, Robert and Rieger, Michael},
  year = 2020,
  month = feb,
  number = {arXiv:1901.07329},
  eprint = {1901.07329},
  primaryclass = {cs},
  publisher = {arXiv},
  doi = {10.48550/arXiv.1901.07329},
  urldate = {2025-04-21},
  archiveprefix = {arXiv}
}

@misc{Jiang_KATS_2022,
  title = {Kats},
  shorttitle = {Kats},
  author = {Jiang, Xiaodong and Srivastava, Sudeep and Chatterjee, Sourav and Yu, Yang and Handler, Jeffrey and Zhang, Peiyi and Bopardikar, Rohan and Li, Dawei and Lin, Yanjun and Thakore, Uttam and Brundage, Michael and Holt, Ginger and Komurlu, Caner and Nagalla, Rakshita and Wang, Zhichao and Sun, Hechao and Gao, Peng and Cheung, Wei and Gao, Jun and Wang, Qi and Guerard, Marius and Kazemi, Morteza and Chen, Yulin and Zhou, Chong and Lee, Sean and Laptev, Nikolay and Levendovszky, Tiham{\'e}r and Taylor, Jake and Qian, Huijun and Zhang, Jian and Shoydokova, Aida and Singh, Trisha and Zhu, Chengjun and Baz, Zeynep and Bergmeir, Christoph and Yu, Di and Koylan, Ahmet and Jiang, Kun and Temiyasathit, Ploy and Yurtbay, Emre},
  year = 2022,
  month = mar,
  copyright = {MIT License}
}

@article{johnsonMIMICIIIFreelyAccessible2016,
  title = {{{MIMIC-III}}, a Freely Accessible Critical Care Database},
  shorttitle = {{{MIMIC-III}}},
  author = {Johnson, Alistair E. W. and Pollard, Tom J. and Shen, Lu and Lehman, Li-wei H. and Feng, Mengling and Ghassemi, Mohammad and Moody, Benjamin and Szolovits, Peter and Anthony Celi, Leo and Mark, Roger G.},
  year = 2016,
  month = may,
  journal = {Scientific Data},
  volume = {3},
  number = {1},
  pages = {160035},
  publisher = {Nature Publishing Group},
  issn = {2052-4463},
  doi = {10.1038/sdata.2016.35},
  urldate = {2022-01-05},
  copyright = {2016 The Author(s)},
  langid = {english}
}

@inproceedings{keLightGBMHighlyEfficient2017,
  title = {{{LightGBM}}: {{A Highly Efficient Gradient Boosting Decision Tree}}},
  shorttitle = {{{LightGBM}}},
  booktitle = {Advances in {{Neural Information Processing Systems}}},
  author = {Ke, Guolin and Meng, Qi and Finley, Thomas and Wang, Taifeng and Chen, Wei and Ma, Weidong and Ye, Qiwei and Liu, Tie-Yan},
  year = 2017,
  volume = {30},
  publisher = {Curran Associates, Inc.},
  urldate = {2024-09-09}
}

@misc{kimMedicalHallucinationsFoundation2025,
  title = {Medical {{Hallucinations}} in {{Foundation Models}} and {{Their Impact}} on {{Healthcare}}},
  author = {Kim, Yubin and Jeong, Hyewon and Chen, Shan and Li, Shuyue Stella and Park, Chanwoo and Lu, Mingyu and Alhamoud, Kumail and Mun, Jimin and Grau, Cristina and Jung, Minseok and Gameiro, Rodrigo and Fan, Lizhou and Park, Eugene and Lin, Tristan and Yoon, Joonsik and Yoon, Wonjin and Sap, Maarten and Tsvetkov, Yulia and Liang, Paul and Xu, Xuhai and Liu, Xin and Park, Chunjong and Lee, Hyeonhoon and Park, Hae Won and McDuff, Daniel and Tulebaev, Samir and Breazeal, Cynthia},
  year = 2025,
  month = nov,
  number = {arXiv:2503.05777},
  eprint = {2503.05777},
  primaryclass = {cs},
  publisher = {arXiv},
  doi = {10.48550/arXiv.2503.05777},
  urldate = {2026-03-13},
  archiveprefix = {arXiv}
}

@misc{koFeRGLLMFeatureEngineering2025,
  title = {{{FeRG-LLM}} : {{Feature Engineering}} by {{Reason Generation Large Language Models}}},
  shorttitle = {{{FeRG-LLM}}},
  author = {Ko, Jeonghyun and Park, Gyeongyun and Lee, Donghoon and Lee, Kyunam},
  year = 2025,
  month = mar,
  number = {arXiv:2503.23371},
  eprint = {2503.23371},
  primaryclass = {cs},
  publisher = {arXiv},
  doi = {10.48550/arXiv.2503.23371},
  urldate = {2025-06-23},
  archiveprefix = {arXiv}
}

@inproceedings{liLearningDataDrivenPolicy2022,
  title = {Learning a {{Data-Driven Policy Network}} for {{Pre-Training Automated Feature Engineering}}},
  shorttitle = {{{FETCH}}},
  booktitle = {The {{Eleventh International Conference}} on {{Learning Representations}}},
  author = {Li, Liyao and Wang, Haobo and Zha, Liangyu and Huang, Qingyi and Wu, Sai and Chen, Gang and Zhao, Junbo},
  year = 2022,
  month = sep,
  urldate = {2025-06-23},
  langid = {english}
}

@article{limTimeseriesForecastingDeep2021,
  title = {Time-Series Forecasting with Deep Learning: A Survey},
  shorttitle = {Time-Series Forecasting with Deep Learning},
  author = {Lim, Bryan and Zohren, Stefan},
  year = 2021,
  month = apr,
  journal = {Philosophical Transactions. Series A, Mathematical, Physical, and Engineering Sciences},
  volume = {379},
  number = {2194},
  pages = {20200209},
  issn = {1471-2962},
  doi = {10.1098/rsta.2020.0209},
  langid = {english},
  pmid = {33583273}
}

@misc{liptonLearningDiagnoseLSTM2017,
  title = {Learning to {{Diagnose}} with {{LSTM Recurrent Neural Networks}}},
  author = {Lipton, Zachary C. and Kale, David C. and Elkan, Charles and Wetzel, Randall},
  year = 2017,
  month = mar,
  number = {arXiv:1511.03677},
  eprint = {1511.03677},
  primaryclass = {cs},
  publisher = {arXiv},
  doi = {10.48550/arXiv.1511.03677},
  urldate = {2026-03-13},
  archiveprefix = {arXiv}
}

@book{littleStatisticalAnalysisMissing2019a,
  title = {Statistical {{Analysis}} with {{Missing Data}}},
  author = {Little, Roderick J. A. and Rubin, Donald B.},
  year = 2019,
  month = apr,
  publisher = {John Wiley \& Sons},
  googlebooks = {BemMDwAAQBAJ},
  isbn = {978-0-470-52679-8},
  langid = {english}
}

@inproceedings{liuControlBurnFeatureSelection2021,
  title = {{{ControlBurn}}: {{Feature Selection}} by {{Sparse Forests}}},
  shorttitle = {{{ControlBurn}}},
  booktitle = {Proceedings of the 27th {{ACM SIGKDD Conference}} on {{Knowledge Discovery}} \& {{Data Mining}}},
  author = {Liu, Brian and Xie, Miaolan and Udell, Madeleine},
  year = 2021,
  month = aug,
  eprint = {2107.00219},
  primaryclass = {cs},
  pages = {1045--1054},
  doi = {10.1145/3447548.3467387},
  urldate = {2026-03-12},
  archiveprefix = {arXiv}
}

@misc{lubbaCatch22CAnonicalTimeseries2019,
  title = {Catch22: {{CAnonical Time-series CHaracteristics}}},
  shorttitle = {Catch22},
  author = {Lubba, Carl H. and Sethi, Sarab S. and Knaute, Philip and Schultz, Simon R. and Fulcher, Ben D. and Jones, Nick S.},
  year = 2019,
  month = jan,
  number = {arXiv:1901.10200},
  eprint = {1901.10200},
  primaryclass = {cs},
  publisher = {arXiv},
  doi = {10.48550/arXiv.1901.10200},
  urldate = {2025-06-23},
  archiveprefix = {arXiv}
}

@misc{lundbergConsistentIndividualizedFeature2019,
  title = {Consistent {{Individualized Feature Attribution}} for {{Tree Ensembles}}},
  author = {Lundberg, Scott M. and Erion, Gabriel G. and Lee, Su-In},
  year = 2019,
  month = mar,
  number = {arXiv:1802.03888},
  eprint = {1802.03888},
  primaryclass = {cs},
  publisher = {arXiv},
  doi = {10.48550/arXiv.1802.03888},
  urldate = {2026-03-12},
  archiveprefix = {arXiv}
}

@article{luSurveyDeepLearning2025,
  title = {A {{Survey}} of {{Deep Learning}} for {{Time Series Forecasting}}: {{Theories}}, {{Datasets}}, and {{State-of-the-Art Techniques}}},
  shorttitle = {A {{Survey}} of {{Deep Learning}} for {{Time Series Forecasting}}},
  author = {Lu, Gaoyong and Ou, Yang and Wang, Zhihong and Qu, Yingnan and Xia, Yingsheng and Tang, Dibin and Kotenko, Igor and Li, Wei},
  year = 2025,
  journal = {Computers, Materials \& Continua},
  volume = {85},
  number = {2},
  pages = {2403--2441},
  publisher = {Tech Science Press},
  issn = {1546-2218, 1546-2226},
  doi = {10.32604/cmc.2025.068024},
  urldate = {2026-03-13},
  langid = {english}
}

@misc{namOptimizedFeatureGeneration2024,
  title = {Optimized {{Feature Generation}} for {{Tabular Data}} via {{LLMs}} with {{Decision Tree Reasoning}}},
  shorttitle = {{{OCTree}}},
  author = {Nam, Jaehyun and Kim, Kyuyoung and Oh, Seunghyuk and Tack, Jihoon and Kim, Jaehyung and Shin, Jinwoo},
  year = 2024,
  month = nov,
  number = {arXiv:2406.08527},
  eprint = {2406.08527},
  primaryclass = {cs},
  publisher = {arXiv},
  doi = {10.48550/arXiv.2406.08527},
  urldate = {2025-02-26},
  archiveprefix = {arXiv}
}

@article{pilgramMagnitudeImpactHallucinations2025,
  title = {Magnitude and {{Impact}} of {{Hallucinations}} in {{Tabular Synthetic Health Data}} on {{Prognostic Machine Learning Models}}: {{Validation Study}}},
  shorttitle = {Magnitude and {{Impact}} of {{Hallucinations}} in {{Tabular Synthetic Health Data}} on {{Prognostic Machine Learning Models}}},
  author = {Pilgram, Lisa and Kababji, Samer El and Liu, Dan and Emam, Khaled El},
  year = 2025,
  month = aug,
  journal = {Journal of Medical Internet Research},
  volume = {27},
  number = {1},
  pages = {e77893},
  publisher = {JMIR Publications Inc., Toronto, Canada},
  doi = {10.2196/77893},
  urldate = {2026-03-13},
  langid = {english}
}

@article{pollardEICUCollaborativeResearch2018,
  title = {The {{eICU Collaborative Research Database}}, a Freely Available Multi-Center Database for Critical Care Research},
  shorttitle = {{{eICU}}},
  author = {Pollard, Tom J. and Johnson, Alistair E. W. and Raffa, Jesse D. and Celi, Leo A. and Mark, Roger G. and Badawi, Omar},
  year = 2018,
  month = sep,
  journal = {Scientific Data},
  volume = {5},
  number = {1},
  pages = {180178},
  publisher = {Nature Publishing Group},
  issn = {2052-4463},
  doi = {10.1038/sdata.2018.178},
  urldate = {2021-08-04},
  copyright = {2018 The Author(s)},
  langid = {english}
}

@article{reynaEarlyPredictionSepsis2020,
  title = {Early {{Prediction}} of {{Sepsis From Clinical Data}}: {{The PhysioNet}}/{{Computing}} in {{Cardiology Challenge}} 2019},
  shorttitle = {P19},
  author = {Reyna, Matthew A. and Josef, Christopher S. and Jeter, Russell and Shashikumar, Supreeth P. and Westover, M. Brandon and Nemati, Shamim and Clifford, Gari D. and Sharma, Ashish},
  year = 2020,
  month = feb,
  journal = {Critical Care Medicine},
  volume = {48},
  number = {2},
  pages = {210--217},
  issn = {0090-3493},
  doi = {10.1097/CCM.0000000000004145},
  urldate = {2021-08-06},
  langid = {american}
}

@article{roustanCliniciansGuideLarge2025,
  title = {The {{Clinicians}}' {{Guide}} to {{Large Language Models}}: {{A General Perspective With}} a {{Focus}} on {{Hallucinations}}},
  shorttitle = {The {{Clinicians}}' {{Guide}} to {{Large Language Models}}},
  author = {Roustan, Dimitri and Bastardot, Fran{\c c}ois},
  year = 2025,
  month = jan,
  journal = {Interactive Journal of Medical Research},
  volume = {14},
  number = {1},
  pages = {e59823},
  publisher = {JMIR Publications Inc., Toronto, Canada},
  doi = {10.2196/59823},
  urldate = {2026-03-13},
  copyright = {Unless stated otherwise, all articles are open-access distributed under the terms of the Creative Commons Attribution License (http://creativecommons.org/licenses/by/3.0/), which permits unrestricted use, distribution, and reproduction in any medium, provided the original work ("first published in the interactive Journal of Medical Research...") is properly cited with original URL and bibliographic citation information. The complete bibliographic information, a link to the original publication on http://www.i-jmr.org/, as well as this copyright and license information must be included.},
  langid = {english}
}

@misc{shuklaMultiTimeAttentionNetworks2021,
  title = {Multi-{{Time Attention Networks}} for {{Irregularly Sampled Time Series}}},
  author = {Shukla, Satya Narayan and Marlin, Benjamin M.},
  year = 2021,
  month = jun,
  number = {arXiv:2101.10318},
  eprint = {2101.10318},
  primaryclass = {cs},
  publisher = {arXiv},
  doi = {10.48550/arXiv.2101.10318},
  urldate = {2026-03-13},
  archiveprefix = {arXiv}
}

@article{silvaPredictingInHospitalMortalitya,
  title = {Predicting {{In-Hospital Mortality}} of {{ICU Patients}}: {{The PhysioNet}}/{{Computing}} in {{Cardiology Challenge}} 2012},
  shorttitle = {P12},
  author = {Silva, Ikaro and Moody, George and Scott, Daniel J and Celi, Leo A and Mark, Roger G},
  pages = {4},
  langid = {english}
}

@article{tangDemocratizingEHRAnalyses2020,
  title = {Democratizing {{EHR}} Analyses with {{FIDDLE}}: A Flexible Data-Driven Preprocessing Pipeline for Structured Clinical Data},
  shorttitle = {Democratizing {{EHR}} Analyses with {{FIDDLE}}},
  author = {Tang, Shengpu and Davarmanesh, Parmida and Song, Yanmeng and Koutra, Danai and Sjoding, Michael W and Wiens, Jenna},
  year = 2020,
  month = dec,
  journal = {Journal of the American Medical Informatics Association},
  volume = {27},
  number = {12},
  pages = {1921--1934},
  issn = {1527-974X},
  doi = {10.1093/jamia/ocaa139},
  urldate = {2021-08-25},
  langid = {english}
}

@inproceedings{wangMIMICExtractDataExtraction2020,
  title = {{{MIMIC-Extract}}: A Data Extraction, Preprocessing, and Representation Pipeline for {{MIMIC-III}}},
  shorttitle = {{{MIMIC-Extract}}},
  booktitle = {Proceedings of the {{ACM Conference}} on {{Health}}, {{Inference}}, and {{Learning}}},
  author = {Wang, Shirly and McDermott, Matthew B. A. and Chauhan, Geeticka and Ghassemi, Marzyeh and Hughes, Michael C. and Naumann, Tristan},
  year = 2020,
  month = apr,
  pages = {222--235},
  publisher = {ACM},
  address = {Toronto Ontario Canada},
  doi = {10.1145/3368555.3384469},
  urldate = {2021-08-25},
  isbn = {978-1-4503-7046-2},
  langid = {english}
}

@misc{yaoReActSynergizingReasoning2023,
  title = {{{ReAct}}: {{Synergizing Reasoning}} and {{Acting}} in {{Language Models}}},
  shorttitle = {{{ReAct}}},
  author = {Yao, Shunyu and Zhao, Jeffrey and Yu, Dian and Du, Nan and Shafran, Izhak and Narasimhan, Karthik and Cao, Yuan},
  year = 2023,
  month = mar,
  number = {arXiv:2210.03629},
  eprint = {2210.03629},
  primaryclass = {cs},
  publisher = {arXiv},
  doi = {10.48550/arXiv.2210.03629},
  urldate = {2026-03-12},
  archiveprefix = {arXiv}
}

@article{yeRoleArtificialIntelligence2024,
  title = {The Role of Artificial Intelligence for the Application of Integrating Electronic Health Records and Patient-Generated Data in Clinical Decision Support},
  author = {Ye, Jiancheng and Woods, Donna and Jordan, Neil and Starren, Justin},
  year = 2024,
  month = may,
  journal = {AMIA Summits on Translational Science Proceedings},
  volume = {2024},
  pages = {459--467},
  issn = {2153-4063},
  urldate = {2026-03-12},
  pmcid = {PMC11141850},
  pmid = {38827061}
}

@misc{zhangMedFeatModelAwareExplainabilityDriven2026,
  title = {{{MedFeat}}: {{Model-Aware}} and {{Explainability-Driven Feature Engineering}} with {{LLMs}} for {{Clinical Tabular Prediction}}},
  shorttitle = {{{MedFeat}}},
  author = {Zhang, Zizheng and Li, Yiming and Xu, Justin and Wang, Jinyu and Wang, Rui and Song, Lei and Bian, Jiang and Eyre, David W. and Fu, Jingjing},
  year = 2026,
  month = feb,
  number = {arXiv:2603.02221},
  eprint = {2603.02221},
  primaryclass = {cs},
  publisher = {arXiv},
  doi = {10.48550/arXiv.2603.02221},
  urldate = {2026-03-12},
  archiveprefix = {arXiv}
}

@misc{zhangOpenFEAutomatedFeature2023,
  title = {{{OpenFE}}: {{Automated Feature Generation}} with {{Expert-level Performance}}},
  shorttitle = {{{OpenFE}}},
  author = {Zhang, Tianping and Zhang, Zheyu and Fan, Zhiyuan and Luo, Haoyan and Liu, Fengyuan and Liu, Qian and Cao, Wei and Li, Jian},
  year = 2023,
  month = jun,
  number = {arXiv:2211.12507},
  eprint = {2211.12507},
  primaryclass = {cs},
  publisher = {arXiv},
  doi = {10.48550/arXiv.2211.12507},
  urldate = {2025-04-21},
  archiveprefix = {arXiv}
}

@article{zhouLargeLanguageModels2025,
  title = {Large Language Models in Biomedicine and Healthcare},
  author = {Zhou, Juexiao and Li, Haoyang and Chen, Siyuan and Chen, Zhangtianyi and Han, Zhongyi and Gao, Xin},
  year = 2025,
  month = dec,
  journal = {npj Artificial Intelligence},
  volume = {1},
  number = {1},
  pages = {44},
  publisher = {Nature Publishing Group},
  issn = {3005-1460},
  doi = {10.1038/s44387-025-00047-1},
  urldate = {2026-03-13},
  copyright = {2025 The Author(s)},
  langid = {english}
}

@article{zhouMissingDataMatter2023,
  title = {Missing Data Matter: An Empirical Evaluation of the Impacts of Missing {{EHR}} Data in Comparative Effectiveness Research},
  shorttitle = {Missing Data Matter},
  author = {Zhou, Yizhao and Shi, Jiasheng and Stein, Ronen and Liu, Xiaokang and Baldassano, Robert N and Forrest, Christopher B and Chen, Yong and Huang, Jing},
  year = 2023,
  month = jul,
  journal = {Journal of the American Medical Informatics Association},
  volume = {30},
  number = {7},
  pages = {1246--1256},
  issn = {1527-974X},
  doi = {10.1093/jamia/ocad066},
  urldate = {2026-03-12}
}
